\RequirePackage{fix-cm}
\documentclass[twocolumn]{svjour3}          % twocolumn
\smartqed  % flush right qed marks, e.g. at end of proof
\usepackage{graphicx}
\usepackage{amssymb}
\usepackage{latexsym}
\usepackage{graphicx}
\usepackage{epstopdf}
\usepackage{caption}
\usepackage{url}
\usepackage{xcolor}
\usepackage{amsmath}
\usepackage{color, colortbl}

\usepackage{makecell}
 \usepackage{here}
\usepackage{lineno}
 \journalname{Signal, Image and Video Processing}
\begin{document}

\title{Efficient Masked Face Recognition Method during the COVID-19 Pandemic}


\author{Walid Hariri}


\institute{Walid Hariri \at
             Labged Laboratory, Computer Science department\\ Badji Mokhtar Annaba University \\
              \email{hariri@labged.net}           %  \\
   %        \and
    %      ... \at
     %         Labged Laboratory, Badji Mokhtar Annaba University \\
			%				\email{farah@labged.net}
}

\date{Received: date / Accepted: date}

\maketitle

\begin{abstract}
The coronavirus disease (COVID-19) is an unparalleled crisis leading to a huge number of casualties and security problems. In order to reduce the spread of coronavirus, people often wear masks to protect themselves. This makes face recognition a very difficult task since certain parts of the face are hidden. A primary focus of researchers during the ongoing coronavirus pandemic is to come up with suggestions to handle this problem through rapid and efficient solutions. In this paper, we propose a reliable method based on occlusion removal and deep learning-based features in order to address the problem of the masked face recognition process. The first step is to remove the masked face region. Next, we apply three pre-trained deep Convolutional Neural Networks (CNN) namely, VGG-16, AlexNet, and ResNet-50, and use them to extract deep features from the obtained regions (mostly eyes and forehead regions). The Bag-of-features paradigm is then applied to the feature maps of the last convolutional layer in order to quantize them and to get a slight representation comparing to the fully connected layer of classical CNN.  Finally, Multilayer Perceptron (MLP) is applied for the classification process. Experimental results on Real-World-Masked-Face-Dataset show high recognition performance compared to other state-of-the-art methods.
\end{abstract}

\section{Introduction}
\label{sec:intro}
The COVID-19 can be spread through contact and contaminated surfaces, therefore, the classical biometric systems based on passwords or fingerprints are not anymore safe. Face recognition is safer without any need to touch any device. Recent studies on coronavirus have proven that wearing a face mask by a healthy and infected population reduces considerably the transmission of this virus. 
However, wearing the mask face causes the following problems: 1) fraudsters and thieves take advantage of the mask, stealing and committing crimes without being identified. 2) community access control and face authentication have become very difficult tasks when a grand part of the face is hidden by a mask. 3) existing face recognition methods are not efficient when wearing a mask which cannot provide the whole face image for description. 4) exposing the nose region is very important in the task of face recognition since it is used for face normalization \cite{peng2011training}, pose correction \cite{lu2005matching}, and face matching \cite{koudelka2005prescreener}. Due to these problems, face masks have significantly challenged existing face recognition methods.

To tackle these problems, we distinguish two different tasks namely: \textit{face mask recognition} and \textit{masked face recognition}. The first one checks whether the person is wearing a mask or no. This can be applied in public places where the mask is compulsory. Masked face recognition, on the other hand, aims to recognize a face with a mask basing on the eyes and the forehead regions. In this paper, we handle the second task using a deep learning-based method.  We use a pre-trained deep learning-based model in order to extract features from the unmasked face regions (out of the mask region). It is worth stating that the occlusions in our case can occur in only one predictable facial region (nose and mouth regions), this can be a good guide to handle this problem efficiently.

The rest of this paper is organized as follows: Section 2 presents the related works. In Section 3 we present the motivation and contribution of the paper. The proposed method is detailed in Section 4. Experimental results are presented in Section 5. Conclusion ends the paper.

\section{Related works}
\label{sec:re_work}

Occlusion is a key limitation of real-world 2D face recognition methods. Generally, it comes out from wearing hats, eyeglasses, masks as well as any other objects that can occlude a part of the face while leaving others unaffected. Thus, wearing a mask is considered the most difficult facial occlusion challenge since it occludes a grand part of the face including the nose. Many approaches have been proposed to handle this problem. We can classify them into three categories namely: local matching approach, restoration approach, and occlusion removal approach.

\textbf{Matching approach:} Aims to compare the similarity between images using a matching process. Generally, the face image is sampled into a number of patches of the same size. Feature extraction is then applied to each patch. Finally, a matching process is applied between probe and gallery faces. The advantage of this approach is that the sampled patches are not overlapped, which avoids the influence of occluded regions on the other informative parts. For example, Martinez et aleix. \cite{martinez2002recognizing} sampled the face region into a fixed number of local patches. matching is then applied for similarity measure.

Other methods detect the keypoints from the face image, instead of local patches. For instance, Weng et al. \cite{weng2016robust} proposed to recognize persons of interest from their partial faces. To accomplish this task, they firstly detected keypoints and extract their textural and geometrical features. Next, point set matching is carried out to match the obtained features. Finally, the similarity of the two faces is obtained through the distance between these two aligned feature sets. Keypoint based matching method is introduced in Duan et al. \cite{duan2018topology}. SIFT keypoint descriptor is applied to select the appropriate keypoints. Gabor ternary pattern and point set matching are then applied to match the local keypoints for partial face recognition. In contrast to the above-mentioned methods based on fixed-size patches matching or keypoints detection, McLaughlin et al. \cite{mclaughlin2016largest} applied the largest matching area at each point of the face image without any sampling. 

\textbf{Restoration approach:} Here, the occluded regions in the probe faces are restored according to the gallery ones. 
For instance, Bagchi et al. \cite{bagchi2014robust} proposed to restore facial occlusions. The detection of the occluded regions is
carried out by thresholding the depth map values of the 3D image. Then the restoration is taken on by Principal Component Analysis (PCA) \cite{wold1987principal}. There are also several approaches that rely on the estimation of the occluded parts. Drira et al. \cite{drira20133d} applied a statistical shape model to predict and restore the partial facial curves. Iterative closest point (ICP) algorithm has been used to remove occluded regions in \cite{gawali20143d}. The restoration is applied using a curve, which uses a statistical estimation of the curves to manage the occluded parts. Partially observed curves are completed by using the curves model produced through the PCA technique.

\textbf{Occlusion removal approach:} In order to avoid a bad reconstruction process, these approaches aim to detect regions found to be occluded in the face image and discard them completely from the feature extraction and classification process. Segmentation based approach is one of the best methods that detect firstly the occluded region part and using only the non-occluded part in the following steps. For instance, Priya and Banu \cite{priya2014occlusion} divided the face image into small local patches. Next, to discard the occluded region, they applied the support vector machine classifier to detect them. Finally, a mean-based weight matrix is used on the non-occluded regions for face recognition. Alyuz et al. \cite{alyuz20133} applied an occlusion removal and restoration. They used the global masked projection to remove the occluded regions. Next, the partial Gappy PCA is applied for the restoration using eigenvectors. 
 
Since the publication of AlexNet architecture in 2012 by Krizhevsky et al. \cite{krizhevsky2012imagenet}, deep CNN have become a common approach in face recognition. It has also been successfully used in face recognition under occlusion variation \cite{almabdy2019deep,hariri2017geometrical,kadhim2023face}. It is seen that the deep learning-based method are founded on the fact that the human visual system automatically ignores the occluded regions and only focuses on the non-occluded ones. For example, Song et al. \cite{song2019occlusion} proposed a mask learning technique in order to discard the feature elements of the masked region for the recognition process.  

Inspired by the high performance of CNN based methods that have strong robustness to illumination, facial expression, and facial occlusion changes, we propose in this paper an occlusion removal approach and deep CNN based model to address the problem of masked face recognition during the COVID-19 pandemic. Motivations and more details about the proposed method are presented in the following sections.

\section{Motivation and contribution of the paper}
\label{sec:contr}

Motivated by the efficiency and the facility of the occlusion removal approaches, we apply this strategy to discard the masked regions.
Experimental results are carried out on Real-world Masked Face Recognition Dataset (RMFRD) and  Simulated Masked Face Recognition Dataset (SMFRD) presented in \cite{wang2020masked}. We start by localizing the mask region. To do so, we apply a cropping filter in order to obtain only the informative regions of the masked face (i.e. forehead and eyes). Next, we describe the selected regions using a pre-trained deep learning model as a feature extractor. This strategy is more suitable in real-world applications comparing to restoration approaches. Recently, some works have applied supervised learning on the missing region to restore them such as in \cite{din2020novel}. This strategy, however, is a difficult and highly time-consuming process.
 
Despite the recent breakthroughs of deep learning architectures in pattern recognition tasks, they need to estimate millions of parameters in the fully connected layers that require powerful hardware with high processing capacity and memory. To address this problem, we present in this paper an efficient quantization based pooling method for face recognition using three pre-trained models. To do so, we only consider the feature maps at the last convolutional layers  (also called channels) using Bag-of-Features (BoF) paradigm.

The basic idea of the classical BoF paradigm is to represent images as orderless sets of local features \cite{hariri2021deep}. To get these sets, the first step is to extract local features from the training images, each feature represents a region from the image. Next, the whole features are quantized to compute a codebook. Test image features are then assigned to the nearest code in the codebook to be represented by a histogram.
In the literature, the BoF paradigm has been largely used for handcrafted feature quantization \cite{lobel2013joint} to accomplish image classification tasks. A comparative study between BoF and deep learning for image classification has been made in Loussaief and Abdelkrim \cite{loussaief2018deep}. To take full advantage of the two techniques, in this paper we can consider BoF as a pooling layer in our trainable convolutional layers. This aims to reduce the number of parameters and makes it possible to classify masked face images.

This deep quantization technique presents many advantages. It ensures a lightweight representation that makes the real-world masked face recognition process a feasible task. Moreover, the masked regions vary from one face to another, which leads to informative images of different sizes. The proposed deep quantization allows classifying images from different sizes in order to handle this issue. Besides, the Deep BoF approach uses a differentiable quantization scheme that enables simultaneous training of both the quantizer and the rest of the network, instead of using fixed quantization merely to minimize the model size \cite{passalis2017learning}. It is worth stating that our proposed method doesn't need to be trained on the mission region after removing the mask. It instead improves the generalization of the face recognition process in the presence of the mask during the pandemic of coronavirus.

\begin{figure}[t]
\centering
\captionsetup{justification=centering}
\includegraphics[width=.50\textwidth]{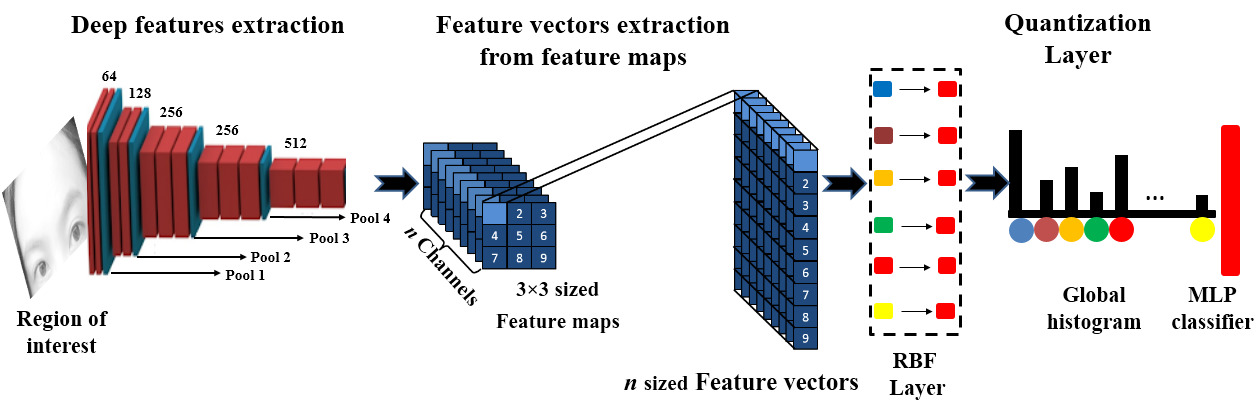}
\caption{Overview of the proposed method.}
\label{fig:overview}
\end{figure}

\begin{figure}[!t]
\centering
\captionsetup{justification=centering}
\includegraphics[width=.50\textwidth]{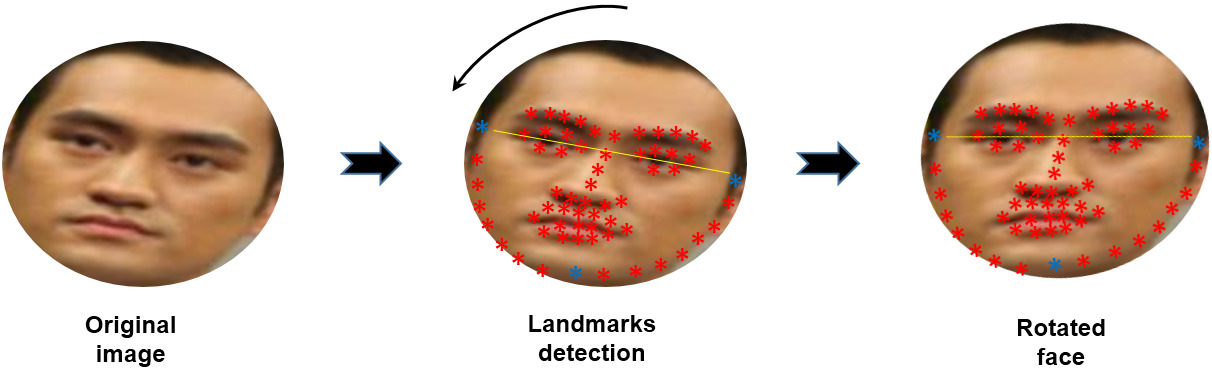}
\caption{2D Face rotation.}
\label{fig:rotation}
\end{figure}

\begin{figure}[!t]
\centering
\captionsetup{justification=centering}
\includegraphics[width=.50\textwidth]{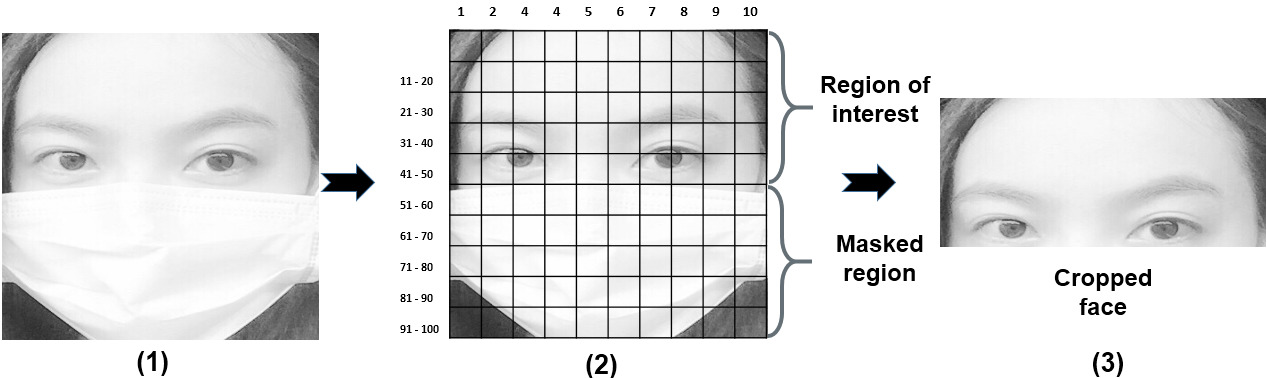}
\caption{(1): Masked face. (2): Sampling the masked face image into 100 regions of the same size. (3): Cropping filter.}
\label{fig:crop}
\end{figure}

\section{The proposed method}
\label{sec:method}

Fig. \ref{fig:overview} presents an overview of the proposed method. It has four steps:

\subsection{Preprocessing and cropping filter}
\label{sec:prep}

The images of the used dataset are already cropped around the face, so we don't need a face detection stage to localize the face from each image. However, we need to correct the rotation of the face so that we can remove the masked region efficiently. To do so, we detect 68 facial landmarks using Dlib-ml open-source library introduced in \cite{king2009dlib}. According to the eye locations, we apply a 2D rotation to make them horizontal as presented in Fig.~\ref{fig:rotation}.

The next step is to apply a cropping filter in order to extract only the non-masked region. To do so, we firstly normalize all face images into 240 $\times$ 240 pixels. Next, we partition a face into blocks. The principle of this technique is to divide the image into 100 fixed-size square blocks (24 $\times$ 24 pixels in our case). Then we extract only the blocks including the non-masked region (blocks from number 1 to 50). Finally, we eliminate the rest of the blocks as presented in Fig.~\ref{fig:crop}.

\subsection{Feature extraction layer}
\label{sec:features}
\begin{figure}[!t]
\centering
\captionsetup{justification=centering}

\includegraphics[width=.40\textwidth]{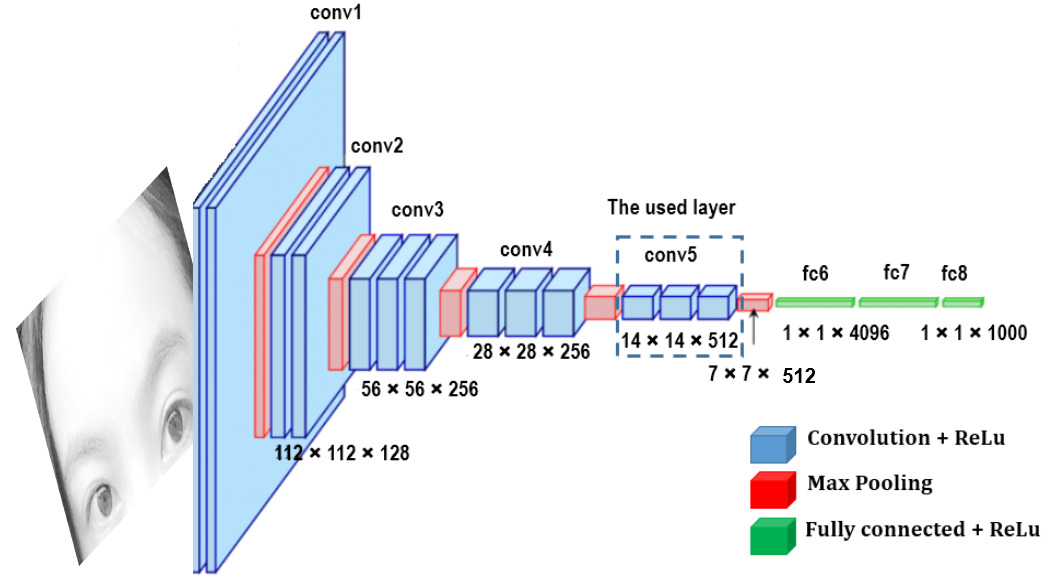}
\caption{VGG-16 network architecture introduced in \cite{simonyan2014very}.}
\label{fig:VGG}
\end{figure}

\begin{figure}[!t]
\centering
\captionsetup{justification=centering}
\includegraphics[width=.45\textwidth]{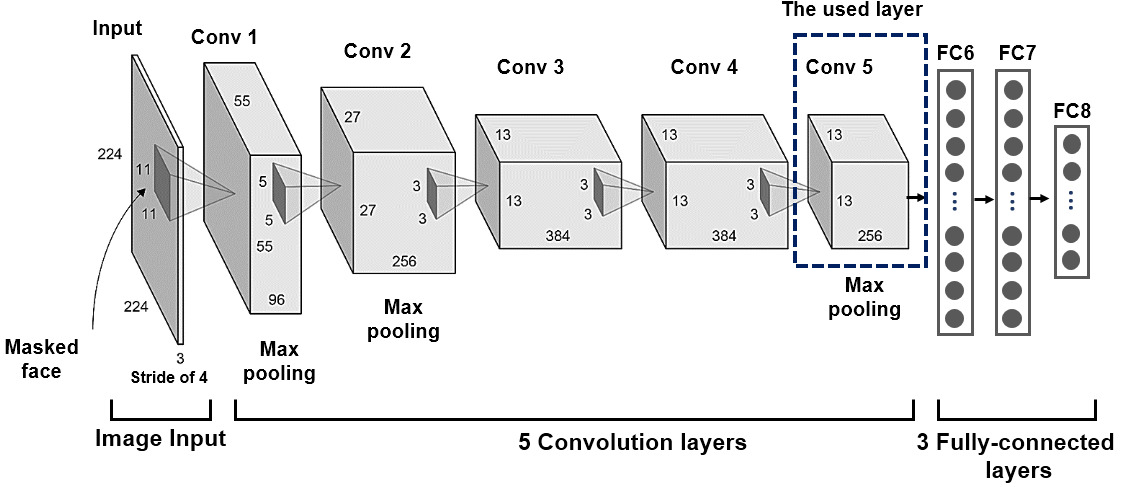}
\caption{AlexNet network architecture introduced in \cite{krizhevsky2017imagenet}.}
\label{fig:alex}
\end{figure}

\begin{figure}[!t]
\centering
\captionsetup{justification=centering}
\includegraphics[width=.50\textwidth]{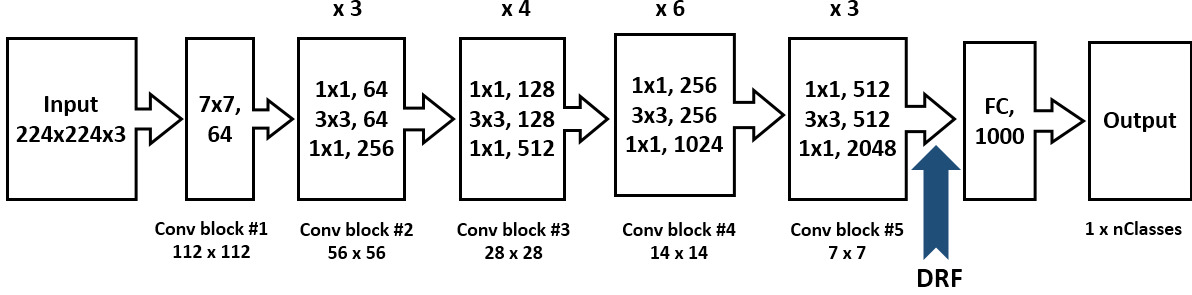}
\caption{ResNet-50 network architecture introduced in \cite{he2016deep}. The extracted DRF are shown.}
\label{fig:resnet50}
\end{figure}

To extract deep features from the informative regions, we have employed three pre-trained models as feature extractors:
\paragraph{\textbf{VGG-16:}} \cite{simonyan2014very} is trained on the ImageNet dataset which has over 14 million images and 1000 classes. Its name VGG-16 comes from the fact that it has 16 layers. It contains different layers including convolutional layers, Max Pooling layers, Activation layers, and Fully Connected (fc) layers. There are 13 convolutional layers, 5 Max Pooling layers, and 3 Dense layers which sum up to 21 layers but only 16 weight layers. In this work, we choose the VGG-16 as the base network, and we only consider the feature maps (FMs) at the last convolutional layer, also called channels. This layer is employed as a feature extractor and will be used for the quantization in the following stage. Fig.~\ref{fig:VGG} presents VGG-16 architecture.
%%%%%%% AlexNet %%%%%%%%%%%
\paragraph{\textbf{AlexNet:}} has been successfully employed for image classification tasks \cite{krizhevsky2017imagenet}. This deep model is pre-trained on a few millions of images from the ImageNet database through eight learned layers, five convolutional layers and three fully-connected layers. The last fully-connected layer allows to classify one thousand classes. The fifth convolutional layer is used in this paper to extract deep features (See Fig.~\ref{fig:alex}). 
%%%%%%%ResNet %%%%%%%%%%%%%%%
\paragraph{\textbf{ResNet-50:}} \cite{he2016deep} has been successfully used in various pattern recognition tasks such as face and pedestrian detection \cite{mliki2020improved}. It containing 50 layers trained on the ImageNet dataset. This network is a combination of Residual network integrations and Deep architecture parsing. Training with ResNet-50 is faster due to the bottleneck blocks. It is composed of five convolutional blocks with shortcuts added between layers. The last convolution layer is used to extract Deep Residual Features (DRF). Fig.~\ref{fig:resnet50} shows the architecture of the ResNet-50 model.

\subsection{Deep bag of features layer}
\label{sec:DB}
From the $i^{th}$ image, we extract feature maps using the feature extraction layer described above. In order to measure the similarity between the extracted feature vectors and the \textit{codewords} also called \textit{term vector}, we applied the RBF kernel as similarity metric as proposed in \cite{passalis2017learning}. Thus, the first sub-layer will be composed of RBF neurons, each neuron is referred to a codeword.

As presented in Fig.~\ref{fig:overview}, the size of the extracted feature map defines the number of the feature vectors that will be used in the BoF layer. Here we refer by $V_i$ to the number of feature vectors extracted from the $i^th$ image. For example, if we have 10 $\times$ 10 feature maps from the last convolutional layer of the chosen pre-trained model, we will have 100 feature vectors to feed the quantization step using the BoF paradigm. To build the \textbf{codebook}, the initialization of the RBF neurons can be carried out manually or automatically using all the extracted feature vectors overall the dataset. The most used automatic algorithm is of course k-means. Let $F$ the set of all the feature vectors, defined by: $F=\{V_{ij}, i=1 \ldots V, j=1 \ldots V_i\}$ and $V_k$ is the number of the RBF neurons centers referred by $c_k$. Note that these RBF centers are learned afterward to get the final codewords. The quantization is then applied to extract the histogram with a predefined number of bins, each bin is referred to a \textit{codeword}. RBF layer is then used as a similarity measure, it contains 2 sub-layers:  

\textbf{(I) RBF layer}: measures the similarity of the input features of the probe faces to the RBF centers. Formally: the $j^{th}$ RBF neuron $\phi(X_j)$ is defined by Eq.(\ref{eq:gauss_kernel}):

\begin{equation}
\phi(X_j)=\exp(\|x-c_j\|_2 /\sigma_j),
\label{eq:gauss_kernel}
\end{equation}

Where $x$ is a feature vector and $c_j$ is the center of the $j^{th}$ RBF neuron. 

\textbf{(II) Quantization layer:} the output of all the RBF neurons is collected in this layer that contains the histogram of the global quantized feature vector that will be used for the classification process. The final histogram is defined by Eq.(\ref{eq:hist}), where $\phi({V})$ is the output vector of the RBF layer over the $c_k$ bins.

\begin{equation}
h_i={V_j}\sum^{N^k}_k \phi({V_{jk}})
\label{eq:hist}
\end{equation}

\subsection{Fully connected layer and classification}
\label{sec:class}

Once the global histogram is computed, we pass to the classification stage to assign each test image to its identity. To do so, we apply the Multilayer perceptron classifier (MLP) where each face is represented by a term vector. Deep BoF network can be trained using back-propagation and gradient descent. Note that the 10-fold cross-validation strategy is applied in our experiments on the RMFRD dataset. We note $V=[v1,\dots,v_k]$ the term vector of each face, where each $v_i$ refers to the occurrence of the term $i$ in the given face. $t$ is the number of attributes, and $m$ is the number of classes (face identities). Test faces are defined by their codeword $V$. MLP uses a set of term occurrences as input values ($v_i$) and associated weights ($w_i$) and a sigmoid function ($g$) that sums the weights and maps the results to output ($y$). Note that the number of hidden layers used in our experiments is given by: $\frac{m+t}{2}$.

\section{Experimental results}
To evaluate the proposed method, we carried out experiments on very challenging masked face datasets. In the following, we present the datasets' content and variations, the experimental results using the quantization of deep features obtained from three pre-trained models, and a comparative study with other state-of-the-arts.

\subsection{Dataset description}
\begin{figure}[h]
\centering
\captionsetup{justification=centering}
\includegraphics[width=.50\textwidth]{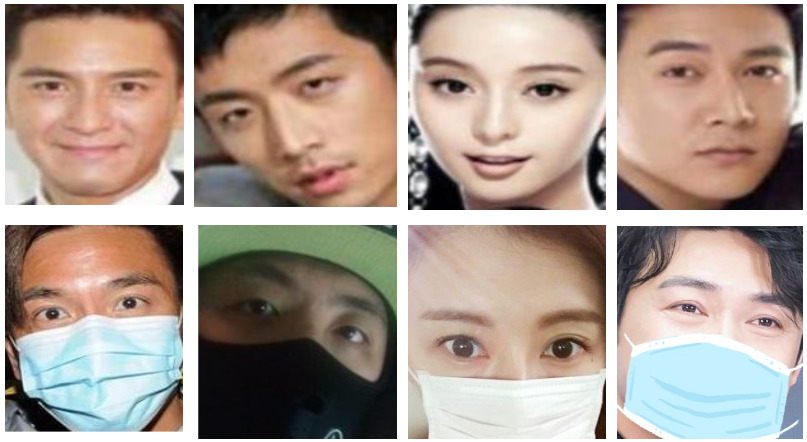}
\caption{Pairs of face images from RMFRD dataset: face images without a mask (up) and with a mask (down).}
\label{fig:RMFRD}
\end{figure}

\begin{figure}[h]
\centering
\captionsetup{justification=centering}
\includegraphics[width=.50\textwidth]{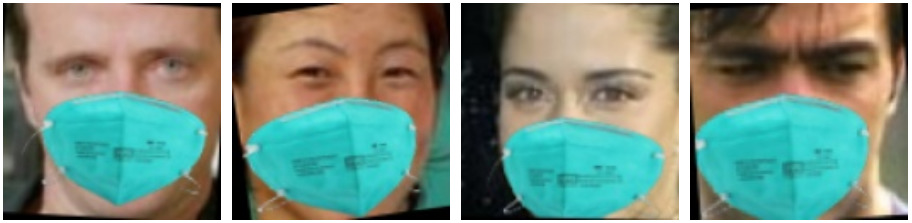}
\caption{Masked faces from SMFRD dataset.}
\label{fig:SMFRD}
\end{figure}

\textbf{Real-World-Masked-Face-Dataset} \cite{wang2020masked} is a masked face dataset devoted mainly to improve the recognition performance of the existing face recognition technology on the masked faces during the COVID-19 pandemic. It contains three types of images namely, Masked Face Detection Dataset (MFDD), Real-world Masked Face Recognition Dataset (RMFRD), and Simulated Masked Face Recognition Dataset (SMFRD). In this paper, we focus on the last two datasets described in the following.
\paragraph {}\textbf{a) RMFRD} is one of the richest real-world masked face datasets. It contains 5,000 images of 525 subjects with masks, and 90,000 images without masks which represent 525 subjects. A semi-automatic annotation strategy has been used to crop the informative face parts. Fig.~\ref{fig:RMFRD} presents some pairs of face images from RMFRD dataset.
\paragraph{}\textbf{b) SMFRD} contains 500,000 simulated masked faces of 10,000 subjects from two known datasets Labeled Faces in the Wild (LFW) \cite{huang2008labeled} and Webface \cite{yi2014learning}. The simulation is carried out using Dlib library \cite{dlibwebsite}. This dataset is balanced but more challenging since the simulated masks are not necessarily in the right position. Fig.~\ref{fig:SMFRD} shows some examples of simulated masked faces from SMFRD dataset.


\subsection{Method performance}
The face images were firstly preprocessed as described in Section \ref{sec:prep}. In contrast to SMFRD dataset, RMFRD is imbalanced (5,000 masked faces vs 90,000 non-masked faces). Therefore, we have applied an over-sampling by cropping some non-masked faces to get an equivalent number of cropped and full faces. Next, using the normalized 2D faces, we employ the three pre-trained models (VGG-16, AlexNet and ResNet-50) separately to extract deep features from their last convolutional layers as presented in Section \ref{sec:features}. The output features are ($14 \times 14 \times 512$, $13 \times 13 \times 256$, $7 \times 7 \times 2048$) dimensional, respectively.  

The quantization is then applied to extract the histogram of a number of bins as presented in Section \ref{sec:DB}. Finally, MLP is applied to classify faces as presented in Section \ref{sec:class}. In this experiment, the 10-fold cross-validation strategy is used to evaluate the recognition performance. The experiments are repeated ten times in RMFRD and SMFRD datasets separately, where 9 samples are used as the training set and the remaining sample as the testing set, and the average results are calculated.

Table \ref{tab:resRMFRD} reports the classification rates on the RMFRD dataset using four different sizes of the codebook (i.e. number of codewords in RBF layer) by (i.e. 50, 60, 70, 100 term vectors per image). We can see that the best recognition rate is obtained using the third FMs in the last convolutional layer from VGG-16 with 60 codewords by 91.3\%. The second FMs achieved 90.8\% with 50 codewords and outperformed the first FMs over the four codeword sizes. AlexNet, on the other hand, realized 86.6\% with 100 codewords where the best recognition rate achieved by ResNet-50 was 89.5\% with 70 codewords. In this experiment, it is clear that VGG-16 outperformed the AlexNet and ResNet-50 models.

Table \ref{tab:resSMFRD} reports the classification rates on the SMFRD dataset. The highest recognition rate is achieved by the ResNet-50 through the quantization of DRF features by 88.9\%. This performance is achieved using 70 codewords that feed an MLP classifier. AlexNet model realized good recognition rates comparing to the VGG-16 model (86.0\% vs 85.6\% as highest rates).
\begin{table}
\centering
\scriptsize
\captionsetup{justification=centering}
\caption{Recognition performance on RMFRD dataset using four codebook sizes.}
\setlength\arrayrulewidth{1pt}
\begin{tabular}{l c c c c}\hline 
\textbf{Method}&\textbf{Size 1}&\textbf{Size 2}&\textbf{Size 3}&\textbf{Size 4}\\  
 term vectors              & 50 &60 &70&100  \\ \hline
&\textbf{VGG-16}&\textbf{model}&& \\  \hline
\makecell{Conv5 FM1 \\ 14$\times$14$\times$512}& 88.5\%&89.2\%&87.1\%&87.5\%\\ \hline 
\makecell{Conv5 FM2 \\ 14$\times$14$\times$512}& 90.8\%&87.4\%&87.2\%&88.0\%\\ \hline      
\makecell{Conv5 FM3 \\ 14$\times$14$\times$512}& 91.0\%&\cellcolor{gray}\textbf{91.3\%}&90.1\%&89.8\%\\ \hline  \hline
&\textbf{AlexNet}&\textbf{model}&& \\\hline     
\makecell{Conv5 FM \\13 $\times$ 13 $\times$ 256}& 84.3\%&85.7\% &85.9\%&86.6\%\\ \hline  \hline
&\textbf{ResNet-50}&\textbf{model}&& \\  \hline     
\makecell{Conv5 FM \\ 7$\times$7$\times$2048}& 87.4\%&87.9\%&89.5\%&89.3\%\\ \hline                    
\end{tabular}
\label{tab:resRMFRD}
\end{table}
%%%
\begin{table}
\centering
\scriptsize
\captionsetup{justification=centering}
\caption{Recognition performance on SMFRD dataset using four codebook sizes.}
\setlength\arrayrulewidth{1pt}
\begin{tabular}{l c c c c}\hline 
\textbf{Method}&\textbf{Size 1}&\textbf{Size 2}&\textbf{Size 3}&\textbf{Size 4}\\  
 term vectors              & 50 &60 &70&100  \\ \hline
&\textbf{VGG-16}&\textbf{model}&& \\  \hline
\makecell{Conv5 FM1 \\ 14$\times$14$\times$512}& 82.4\%&83.7\%&84.5\%&84.7\%\\ \hline 
\makecell{Conv5 FM2 \\ 14$\times$14$\times$512}& 83.1\%&83.5\%&85.0\%&85.4\%\\ \hline      
\makecell{Conv5 FM3 \\ 14$\times$14$\times$512}& 81.7\%&81.3\%&84.4\%&85.6\\ \hline  \hline
&\textbf{AlexNet}&\textbf{model}&& \\\hline     
\makecell{Conv5 FM \\13 $\times$ 13 $\times$ 256}& 83.7\%&83.9\% &84.2\%&86.0\%\\ \hline  \hline
&\textbf{ResNet-50}&\textbf{model}&& \\  \hline     
\makecell{Conv5 FM \\ 7$\times$7$\times$2048}& 83.5\%&84.7\%&\cellcolor{gray}\textbf{88.9\%}&88.5\%\\ \hline                    
\end{tabular}
\label{tab:resSMFRD}
\end{table}

\subsection{Performance comparison}
To further evaluate the performance of our proposed method, we have compared the obtained experimental results with those of other face recognizers on the RMFRD and SMFRD datasets as follows:

\paragraph{\textbf{Comparison with transfer learning-based technique:}} We have tested the face recognizer presented in \cite{luttrell2018deep} that achieved a good recognition accuracy on two subsets of the FERET database \cite{phillips1998feret}. This technique is based on transfer learning (TL) which employs pre-trained models and fine-tuning them to recognize masked faces from RMFRD and SMFRD datasets. The reported results in Table \ref{tab:compari_recognizer} show that the proposed method outperformed the TL-based method on the RMFRD and SMFRD datasets.

\paragraph{\textbf{Comparison with covariance-based technique:}} Covariance-based features have been applied in \cite{hariri20163d} and achieved high recognition performance on 3D datasets in the presence of occluded regions. We have employed this method using 2D-based features (texture, gray level, LBP) to extract covariance descriptors. The evaluation on the RMFRD and SMFRD datasets confirms the superiority of the proposed method as shown in Table \ref{tab:compari_recognizer}.
\paragraph{\textbf{Comparison with deep feature extractor:}} Another efficient face recognition method using the same pre-trained models (AlexNet and ResNet-50) is proposed in \cite{almabdy2019deep} and achieved a high recognition rate on various datasets. Nevertheless, the pre-trained models are employed in a different manner. It consists of applying a TL technique to fine-tune the pre-trained models to the problem of masked face recognition using an SVM classifier. We have tested the this strategy on the masked faces. The results in Table \ref{tab:compari_recognizer} further demonstrate the efficiency of the BoF paradigm compared to the use of a machine learning-based classifier directly.
\begin{table}[h]
\centering
\scriptsize
\captionsetup{justification=centering}
\caption{Performance comparison with state-of-the-art methods.}
\setlength\arrayrulewidth{1pt}
\begin{tabular}{l c c c c}\hline                            
\cellcolor{lightgray} Method&\cellcolor{lightgray}Dataset&\cellcolor{lightgray}Technique&\cellcolor{lightgray}Masks&\cellcolor{lightgray}Accuracy\\ \hline 
Luttrell et al. \cite{luttrell2018deep}&RMFRD&TL   & yes & 85.7\%\\ \hline
Hariri et al. \cite{hariri20163d}&RMFRD&Covariance   & yes & 84.6\%\\ \hline
Almabdy et al. \cite{almabdy2019deep}&RMFRD&CNN+SVM   & yes & 87.0\%\\ \hline
\textbf{Our}&  RMFRD&CNN+BoF  & yes & \cellcolor{gray}\textbf{91.3\%}\\ \hline\hline
Luttrell et al. \cite{luttrell2018deep}&SMFRD&TL  & yes & 83.3\%\\ \hline 
Hariri et al. \cite{hariri20163d}&SMFRD&Covariance   & yes & 83.8\%\\ \hline
Almabdy et al. \cite{almabdy2019deep}&SMFRD&CNN+SVM   & yes & 86.1\%\\ \hline
\textbf{Our}& SMFRD & CNN+BoF& yes & \cellcolor{gray}\textbf{88.9\%}\\ \hline                                
\end{tabular}
\label{tab:compari_recognizer}
\end{table}

\subsection{Computation and training times comparison}
The comparison of the computation times between the proposed method and Almabdy et al.'s method \cite{almabdy2019deep} shows that the use of the BoF paradigm decreases the time required to extract deep features and to classify the masked faces (See Table \ref{tab:time}). Note that this comparison is performed using the same pre-trained models (VGG-16 and AlexNet) on the RMFRD dataset. AlexNet is lowest training and testing time compared to VGG-16 with less GPU memory usage.
\begin{table}[h]
\centering
\scriptsize
\captionsetup{justification=centering}
\caption{Training and testing time on the RMFRD dataset in milliseconds.}
\setlength\arrayrulewidth{1pt}
\begin{tabular}{l c c}\hline                            
\cellcolor{lightgray} Method&\cellcolor{lightgray}AlexNet&\cellcolor{lightgray}VGG-16\\ \hline 
Almabdy et al. \cite{almabdy2019deep}&\makecell{train:550\\test:34}  & \makecell{train:930\\test:120} \\ \hline   
\textbf{Our}&\makecell{train: 308\\test:21}  & \makecell{Train:605\\test:84}   \\ \hline                                       
\end{tabular}
\label{tab:time}
\end{table}

\subsection{Discussion}
The obtained high accuracy compared to other face recognizers is achieved due to the best features extracted from the last convolutional layers of the pre-trained models, and the high efficiency of the proposed BoF paradigm that gives a lightweight and more discriminative power comparing to classical CNN with softmax function. Moreover, dealing with only the unmasked regions, the high generalization of the proposed method makes it applicable in real-time applications. Other methods, however, aim to unmask the masked face using generative networks such as in \cite{din2020novel}. This strategy is a greedy task and not preferable for real-world applications. 

The efficiency of each pre-trained model depends on its architecture and the abstraction level of the extracted features. When dealing with real masked faces, VGG-16 has achieved the best recognition rate, while ResNet-50 outperformed both VGG-16 and AlexNet on the simulated masked faces. This behavior can be explained by the fact that VGG-16 features fail to ensure a high discriminative power comparing to the DRF features that are still relatively steady compared to their results on the real masked faces. When dealing with other state-of-the-art recognizers, one of them applied the same pre-trained models with a different strategy. The proposed method outperformed TL-based method using the same pre-trained models. This performance is explained by the fact that the fc layers of the pre-trained models are more dataset-specific features (generally pre-trained on ImageNet dataset) which is a very different dataset, thus, this strategy is not always suitable for our task. Moreover, the proposed method outperformed previous methods in terms of training time. The achieved performance further confirms that the BoF paradigm is a slight representation that further reinforces the high discrimination power of the deep features to feed a machine learning-based classifier.

\section{Conclusion}

The proposed method improves the generalization of the face recognition process in the presence of the mask. To accomplish this task, we proposed a deep learning-based method and quantization-based technique to deal with the recognition of the masked faces. We employed three pre-trained models and used their last convolutional layers as deep features. The Bof paradigm is applied to quantize the obtained features and to feed an MLP classifier. The proposed method outperformed other state-of-the-art methods in terms of accuracy and time complexity. The proposed method can also be extended to richer applications such as video retrieval and surveillance.  In future work, we look at the application of deep ensemble models with additional pre-trained models to enhance the accuracy.

\bibliographystyle{abbrv}
\bibliography{refs}

\end{document}